%% file: Thesis.tex
\documentclass[sigconf,table,xcdraw]{acmart}

\usepackage{subcaption}
\usepackage{amsmath}
\usepackage{amsfonts}
\usepackage{booktabs}       % professional-quality tables
\usepackage{diagbox}
\usepackage{multirow}
\usepackage[normalem]{ulem}
\useunder{\uline}{\ul}{}
\usepackage{hhline} 
\usepackage{graphicx}

\AtBeginDocument{%
  }

\copyrightyear{2025}
\acmYear{2025}
\setcopyright{acmlicensed}
\acmConference[CIKM '25]{Proceedings of the 34th ACM International Conference on Information and Knowledge Management}{November 10--14, 2025}{Seoul, Republic of Korea}
\acmBooktitle{Proceedings of the 34th ACM International Conference on Information and Knowledge Management (CIKM '25), November 10--14, 2025, Seoul, Republic of Korea}
\acmDOI{10.1145/3746252.3761142}
\acmISBN{979-8-4007-2040-6/2025/11}

\begin{document}

%%
%% The "title" command has an optional parameter,
%% allowing the author to define a "short title" to be used in page headers.
\title{PromptTSS: A Prompting-Based Approach for Interactive Multi-Granularity Time Series Segmentation}

\settopmatter{authorsperrow=4} % compact author block

\author{Ching Chang}
\authornote{Correspondence to: Ching Chang <\texttt{blacksnail789521.cs10@nycu.edu.tw}>}
\affiliation{%
  \department{Computer Science}
  \institution{National Yang Ming Chiao Tung University}
  \city{Hsinchu}
  \country{Taiwan}
}

\additionalaffiliation{%
  \institution{GoEdge.ai}
  \country{Taiwan}
}

\author{Ming-Chih Lo}
\affiliation{%
  \department{Computer Science}
  \institution{National Yang Ming Chiao Tung University}
  \city{Hsinchu}
  \country{Taiwan}
}

\author{Wen-Chih Peng}
\affiliation{%
  \department{Computer Science}
  \institution{National Yang Ming Chiao Tung University}
  \city{Hsinchu}
  \country{Taiwan}
}

\author{Tien-Fu Chen}
\affiliation{%
  \department{Computer Science}
  \institution{National Yang Ming Chiao Tung University}
  \city{Hsinchu}
  \country{Taiwan}
}

%%
%% By default, the full list of authors will be used in the page
%% headers. Often, this list is too long, and will overlap
%% other information printed in the page headers. This command allows
%% the author to define a more concise list
%% of authors' names for this purpose.
\renewcommand{\shortauthors}{Ching et al.}

%%
%% The abstract is a short summary of the work to be presented in the
%% article.
\begin{abstract}
Multivariate time series data, collected across various fields such as manufacturing and wearable technology, exhibit states at multiple levels of granularity, from coarse-grained system behaviors to fine-grained, detailed events.
Effectively segmenting and integrating states across these different granularities is crucial for tasks like predictive maintenance and performance optimization.
However, existing time series segmentation methods face two key challenges: (1) the inability to handle multiple levels of granularity within a unified model, and (2) limited adaptability to new, evolving patterns in dynamic environments. 
To address these challenges, we propose PromptTSS, a novel framework for time series segmentation with multi-granularity states. 
PromptTSS uses a unified model with a prompting mechanism that leverages label and boundary information to guide segmentation, capturing both coarse- and fine-grained patterns while adapting dynamically to unseen patterns.
Experiments show PromptTSS improves accuracy by 24.49\% in multi-granularity segmentation, 17.88\% in single-granularity segmentation, and up to 599.24\% in transfer learning, demonstrating its adaptability to hierarchical states and evolving time series dynamics.
Our code is available at \url{https://github.com/blacksnail789521/PromptTSS}.
\end{abstract}

%%
%% The code below is generated by the tool at http://dl.acm.org/ccs.cfm.
%% Please copy and paste the code instead of the example below.
%%


%% Keywords. The author(s) should pick words that accurately describe
%% the work being presented. Separate the keywords with commas.

% \received{20 February 2007}
% \received[revised]{12 March 2009}
% \received[accepted]{5 June 2009}

%%
%% This command processes the author and affiliation and title
%% information and builds the first part of the formatted document.
\maketitle

\section{Introduction}
\label{sec:introduction}
Multivariate time series data is common in domains such as manufacturing, healthcare, and activity monitoring.
These sequences often contain informative \textit{states} or \textit{segments} that reflect system conditions or human behaviors, which are critical for downstream tasks like predictive maintenance and clinical assessment \cite{prectime, wisdm, ticc, time2state}.

In real-world scenarios, such states occur at multiple levels of granularity.
For example, manufacturing processes may involve both high-level production phases and fine-grained machine operations \cite{hecsl, industrial_multiple_gran}, while healthcare applications track general activities like walking alongside micro-events such as tremors \cite{act_rec_tut, pamap2}.
Capturing both coarse and fine-grained patterns is essential, yet most segmentation models operate at a single granularity level, limiting their practical utility.

% Furthermore, segmentation often requires adaptation at inference time.
% In interactive labeling tools, users may only provide sparse corrections or partial annotations, making retraining infeasible.
% This demands models that can dynamically adjust to user input and support different granularity preferences \cite{non_stationary_industrial, seqsleepnet}.
% Yet current methods fall short in two critical ways: they cannot handle multiple state granularities within a single model, and they lack the flexibility to adapt in real time to new patterns.

Furthermore, segmentation often requires adaptation at inference time.  
In interactive labeling tools, users may only provide sparse corrections or partial annotations, making retraining infeasible.  
This demands models that can dynamically adjust to user input and support different granularity preferences \cite{non_stationary_industrial, seqsleepnet}.  
% A representative case arises in industrial environments, where labeling costs are high, and a model trained on one system must be transferred to a related but distinct system with minimal additional supervision.  
In industrial settings, where labeling is costly, a model trained on one system often needs to transfer to a related system with minimal supervision.
Yet current methods fall short in two critical ways: they cannot handle multiple state granularities within a single model, and they lack the flexibility to adapt in real time to new patterns.

% Challenge 1: Inability to Handle Multi-Granularity
\textbf{Challenge 1: Inability to Handle Multi-Granularity.}
Existing time series segmentation (TSS) models are designed to work at a single level of granularity, which significantly limits their capacity to detect both broad, long-term trends and short, localized events simultaneously \cite{seqsleepnet, ms_tcn, u_time}.
This single-granularity segmentation approach results in models that either overlook fine-grained details or fail to capture broader patterns, making it challenging to effectively capture the full spectrum of events and trends.
For example, in a manufacturing factory, coarse-level segmentation might miss short-lived anomalies like abrupt spikes in temperature at a specific station, while fine-grained segmentation could over-segment data, failing to capture larger patterns such as gradual declines in machine performance across multiple stages.

% Challenge 2: Limited Adaptability to New Patterns
\textbf{Challenge 2: Limited Adaptability to New Patterns.}
Most segmentation models are static, meaning they are trained on specific datasets and lack the ability to adapt to evolving patterns \cite{prectime, deepconvlstm}.
In dynamic environments, such as smart manufacturing or financial markets, time series patterns can change over time, necessitating continuous adaptation \cite{non_stationary_industrial}.
Models trained on historical data often struggle to generalize well to new, unseen behaviors, leading to outdated segmentation rules.
This lack of flexibility hinders the model's ability to accurately detect novel trends or anomalies, ultimately reducing the effectiveness of the analysis in environments where conditions are constantly shifting.
 
\begin{figure}[t]
    \centering
    \includegraphics[width=1.0\columnwidth]{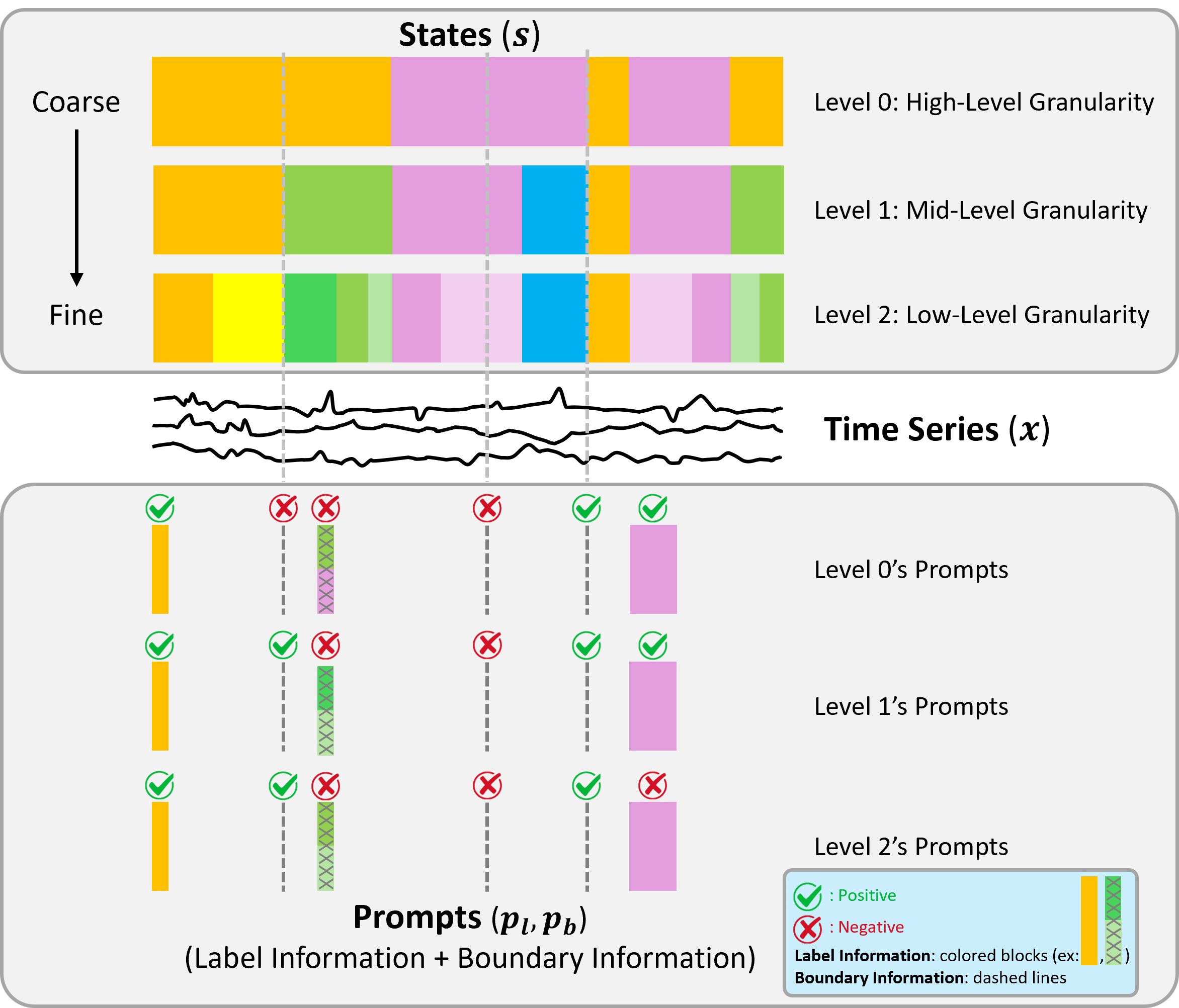}
    \caption{Multiple-Granularity States and Prompts. The top illustrates coarse-to-fine state granularity in time series. The bottom shows two types of prompts in PromptTSS—label and boundary—using green checks for positive information and red crosses for negative information.}
    \label{fig:multi_granuarlities}
\end{figure}

% Our method
To address these challenges, we propose PromptTSS, a unified model for time series segmentation with multi-granularity states, utilizing a novel prompting mechanism guided by a small portion of label and boundary information.
To tackle the first challenge, the prompting mechanism allows the model to select the appropriate granularity, all while utilizing a single unified model. 
This approach enhances the model's ability to capture both coarse-level trends and fine-level events simultaneously. 
For the second challenge, the prompting mechanism enables the model to dynamically adapt to unseen patterns by allowing users to provide prompts during inference.
This capability ensures that the model can continuously incorporate ground truth information, guiding its predictions even in the presence of new or evolving behaviors.
In PromptTSS, two types of ground truth information are used as prompts (as shown in the bottom side of Figure \ref{fig:multi_granuarlities}): label information, which provides contextual state annotations, and boundary information, which indicates the transition points between different states. 
The framework employs a time series encoder, a prompt encoder, and a state decoder. 
The time series encoder and prompt encoder encode time series data and prompts, while the state decoder predicts the corresponding states based on the encoded representations.

In summary, the paper's main contributions are as follows:
\begin{itemize}
    \item \textbf{Unified Multi-Granularity Segmentation}: We propose PromptTSS, the first unified model for segmenting multivariate time series data with multiple granularity levels of states, addressing the challenges of capturing both broad trends and fine-grained events.
    \item \textbf{Dynamic Adaptability with Prompts}: We introduce a novel prompting mechanism that incorporates ground truth information (label and boundary), enabling the model to dynamically adapt to unseen patterns and evolving behaviors in dynamic environments.
    \item \textbf{Comprehensive Validation}: Experiments show PromptTSS achieves 24.49\% and 17.88\% accuracy improvements in multi- and single-granularity segmentation, with up to 599.24\% improvement in transfer learning, showing its strength in multi-granularity modeling and adaptation to unseen data.
\end{itemize}

\section{Related Work}
\label{sec:related_work}

\subsection{Time Series Segmentation}
\label{sec:related_work_tss}
Time series segmentation is generally categorized into change point detection and state detection.
Change point detection identifies transitions in the signal but lacks semantic interpretation for each segment \cite{clasp, floss, change_point_detection_1, change_point_detection_2}.
State detection, which we focus on in this work, assigns a label to each time step, producing interpretable segmentations suitable for downstream tasks \cite{prectime, time2state}.

Methods can be broadly divided into unsupervised and supervised approaches.
Unsupervised methods \cite{time2state, ticc, hvgh, e2usd} use heuristics, clustering, or statistical models without labeled data, but often struggle with low accuracy, limited interpretability, and difficulty modeling complex or hierarchical transitions.
Supervised approaches leverage labeled data to achieve higher accuracy and robustness.
PrecTime~\cite{prectime} combines sliding windows with dense labeling for precise industrial segmentation.
MS-TCN++~\cite{ms_tcn} uses multi-stage temporal convolutions to reduce over-segmentation.
U-Time~\cite{u_time} applies a U-Net-style architecture to 1D time series for local and global context modeling.
DeepConvLSTM~\cite{deepconvlstm} stacks CNNs and LSTMs to capture both spatial and temporal dependencies.
However, these models are typically trained for a fixed granularity level and lack support for user interaction or real-time adaptation.
In contrast, our work introduces a prompting-based model that enables interactive segmentation across multiple granularities in a unified framework.

\subsection{Connections to Broader ML Paradigms in Time Series Segmentation}
While PromptTSS is designed for interactive multi-granularity time series segmentation, it relates conceptually to several machine learning paradigms, including \textit{active learning}, \textit{domain adaptation}, \textit{hierarchical classification}, and \textit{multi-label learning}.  
Below, we outline these connections and highlight the distinct role PromptTSS plays.
\textit{Active learning} aims to identify informative samples for annotation to improve efficiency \cite{active_1, active_2, active_3}, but typically requires retraining, which is impractical for real-time segmentation.  
PromptTSS allows users to inject sparse prompts during inference, enabling immediate adaptation in dynamic or streaming environments.
\textit{Domain adaptation and distribution shift} are often addressed through fine-tuning or domain-invariant learning \cite{adaption_1, adaption_2, adaption_3}.  
PromptTSS instead uses prompting as a lightweight, human-in-the-loop mechanism to guide inference on unfamiliar or evolving time series.
\textit{Hierarchical classification} enforces consistency across label hierarchies, usually in static data \cite{coke, hierarchical_1, hierarchical_2}.  
PromptTSS extends this to time series, enabling users to control granularity through prompts and refine state structures interactively.
\textit{Multi-label learning} generally assumes fixed label sets on pre-segmented inputs \cite{multi_label_1, multi_label_2, multi_label_3}.  
In contrast, PromptTSS supports sparse, multi-scale supervision by allowing users to clarify label combinations and boundaries during inference.
In summary, PromptTSS draws inspiration from these paradigms but unifies them into a real-time, prompt-guided framework for time series segmentation that addresses limitations none of them handle individually.

\subsection{Prompting}
\label{sec:related_work_prompting}
Prompting has recently emerged as an effective mechanism for enhancing the adaptability of machine learning models, particularly in natural language processing (NLP) and computer vision (CV).
In NLP, large language models (LLMs) \cite{chain_of_though, nlp_prompting_survey} leverage prompts to guide their responses and adapt to new tasks without fine-tuning.
Similarly, in CV, recent advancements such as Segment Anything Model (SAM) \cite{sam} and other works \cite{cv_interative_1, cv_interative_2} utilize prompt-based interactions to enhance segmentation performance and enable user-controlled refinements.
These approaches demonstrate that prompting can significantly improve model adaptability and interaction, allowing models to incorporate external guidance effectively.

While prompting has been extensively studied in NLP and CV, its application to time series analysis remains underexplored.
Most existing time series segmentation models operate in a static manner, lacking mechanisms to integrate user guidance dynamically.
To bridge this gap, our work introduces a prompting mechanism tailored for time series segmentation, enabling interactive and multi-granularity state detection.

\begin{figure*}[t]
    \centering
    \includegraphics[width=1.8\columnwidth]{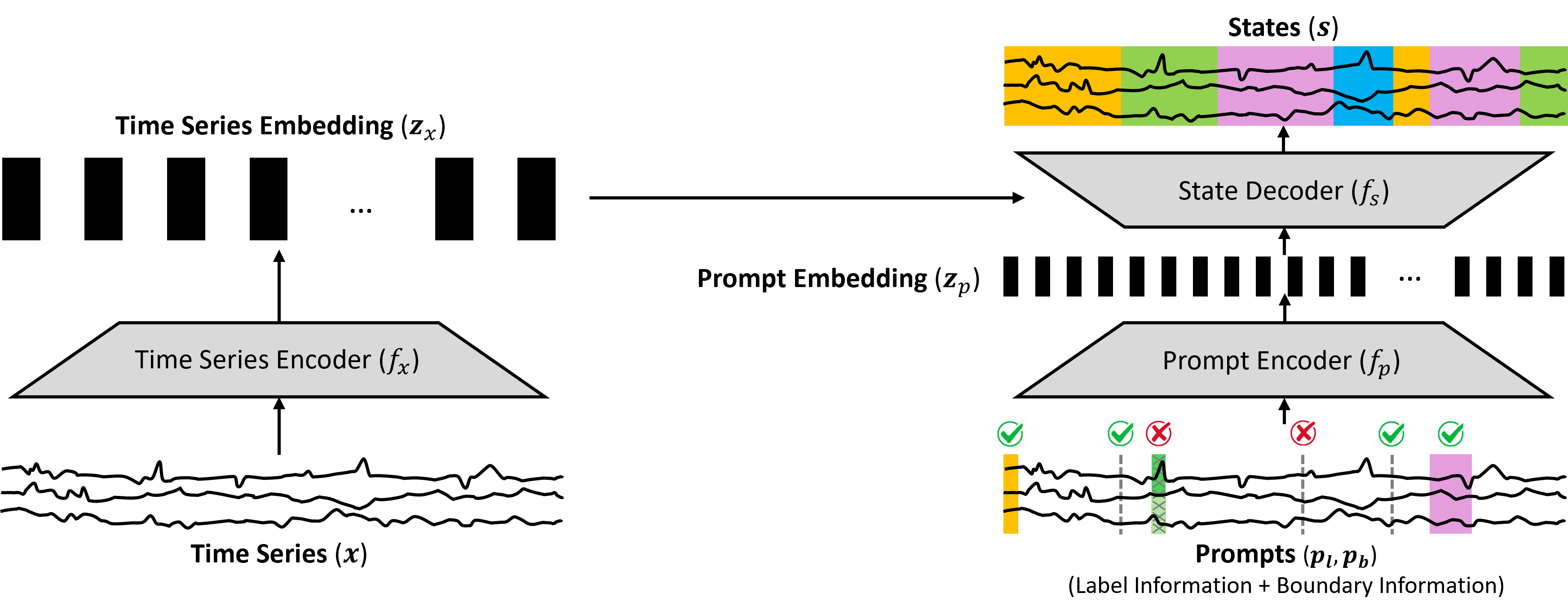}
    \caption{\textbf{PromptTSS framework}. Consists of three modules: a time series encoder that generates time series embeddings, a prompt encoder that produces prompt embeddings, and a state decoder that predicts states based on both embeddings.}
    \label{fig:framework}
\end{figure*}

\section{Problem Formulation}
\label{sec:problem_formulation}
Given a complete and evenly-sampled multivariate time series and its corresponding sequence of discrete states, we use a sliding window of length \(T\) and stride \(S\) to extract sequential samples.
Each sliding window captures a sequence of the time series data, denoted as \(\mathbf{x} = (x_1, \ldots, x_T) \in \mathbb{R}^{T \times C}\), where \(x_t\) is a \(C\)-dimensional vector representing the features at time step \(t\).
The corresponding sequence of states is denoted as \(\mathbf{s} = (s_1, \ldots, s_T) \in \mathbb{Z}^T\), where each \(s_t\) belongs to one of \(K\) discrete states.

In addition to the time series data, auxiliary prompts are introduced to guide the segmentation process.
The prompts consist of label prompts \(\mathbf{p}_l = (p_{l,1}, \ldots, p_{l,T}) \in \mathbb{R}^{T \times D_l}\), which provide guidance on state assignments for each timestamp, and boundary prompts \(\mathbf{p}_b = (p_{b,1}, \ldots, p_{b,T}) \in \mathbb{R}^{T \times D_b}\), which encode information about state transitions.
Here, \( D_l \) and \( D_b \) represent the dimensions of the label and boundary prompt features, respectively.
The objective is to use the input data \(\mathbf{x}\) along with the prompts \(\mathbf{p}_l\) and \(\mathbf{p}_b\) to predict the corresponding sequence of discrete states \(\mathbf{s}\).

\section{Method}
\label{sec:method}
Figure \ref{fig:framework} shows the PromptTSS architecture, which consists of three modules.
The time series encoder encodes time series data into time series embeddings (Section \ref{sec:time_series_encoder}), the prompt encoder processes prompts into prompt embeddings (Section \ref{sec:prompt_encoder}), and the state decoder predicts the corresponding states based on both time series embeddings and prompt embeddings (Section \ref{sec:state_decoder}).

\subsection{Time Series Encoder}
\label{sec:time_series_encoder}
Given time series data \(\mathbf{x}\), the Transformer encoder (\(f_x\)) is used as the time series encoder to generate embeddings \(\mathbf{z}_x\).
Due to the lengthy nature of time series data, directly feeding it into a Transformer encoder can be computationally expensive.
To address this issue, \textit{patching} \cite{patchtst} is employed, where adjacent time steps are aggregated into a single patch-based token to reduce the computational burden.
Patching has been shown to be effective not only in the time series domain but also in other fields such as computer vision \cite{sam} and audio processing \cite{ast}. 

The process begins with patching the time series:
\begin{equation}
\mathbf{x}_{\text{patched}} = \text{Patching}(\mathbf{x}),
\end{equation}
where \( \text{Patching}(\cdot) \) aggregates adjacent time steps into the patched time series \( \mathbf{x}_{\text{patched}} \in \mathbb{R}^{T_{\text{patched}} \times (C \cdot P)}\). Here, \( T_{\text{patched}} \) denotes the number of patch tokens, and \( P \) is the patch length.
The Transformer encoder \( f_x \) is then applied to the patched time series:
\begin{equation}
\mathbf{z}_x = f_x(\mathbf{x}_{\text{patched}}),
\end{equation}
where \( \mathbf{z}_x \in \mathbb{R}^{T_{\text{patched}} \times D} \) represents the time series embeddings, with \( D \) being the embedding dimension.

\subsection{Prompt Encoder}
\label{sec:prompt_encoder}

A novel prompting mechanism is introduced, enabling users to specify a desired level of granularity during inference or guide the model in adapting to unseen patterns.
The prompt encoder (\( f_p \)) is designed to transform the provided prompts into embeddings \(\mathbf{z}_p\) that effectively guide the model.
In PromptTSS, two types of prompts are utilized: label prompts (\( \mathbf{p}_l \)) and boundary prompts (\( \mathbf{p}_b \)), each capable of incorporating both positive and negative information.

\subsubsection{Label Prompts}
\label{sec:label_prompts}
For label prompts, the sequence of prompts is denoted as \( \mathbf{p}_l = (p_{l,1}, \ldots, p_{l,T}) \).
The positive information in label prompts provides a single state for a given timestamp, guiding the model toward the target state.
The negative information in label prompts indicates states that should not be predicted for a given timestamp, helping the model refine its assignments.
Notably, for label prompts, both positive and negative information can be provided simultaneously for the same timestamp.
However, each given timestamp can be assigned only one piece of positive information, while multiple pieces of negative information may be provided to refine the model’s predictions.

To encode label prompts, the positive and negative information for each timestamp are first collected to form a single vector \( \mathbf{p}_{l,t} \in \{0, 1\}^{2K} \), where \( K \) represents the total number of possible states. 
Specifically, the first \( K \)-dimensional space represents the positive label using a one-hot encoding (only one element can be 1, while all others are 0), while the second \( K \)-dimensional space represents the negative labels using a multi-hot encoding (multiple elements can be 1 to indicate the states that should not be predicted). 
The vector \( \mathbf{p}_{l,t} \) is then passed through a linear layer \(\text{Linear}_l\) to transform its dimension into the embedding space of size \( D \):
\begin{equation}
\mathbf{e}_{l,t} = \text{Linear}_l(\mathbf{p}_{l,t}),
\end{equation}
where \( \mathbf{e}_{l,t} \in \mathbb{R}^D\) denotes the embedding of the label prompt for timestamp \( t \).

\subsubsection{Boundary Prompts}
\label{sec:boundary_prompts}
For boundary prompts, the sequence of prompts is denoted as \( \mathbf{p}_b = (p_{b,1}, \ldots, p_{b,T}) \), where \( T \) is the total number of timestamps. 
The positive information in boundary prompts indicates that a state transition should occur at a specific timestamp, while the negative information signals that a state transition should not occur, thereby maintaining the current state. 
Unlike label prompts, boundary prompts allow only one type of information—either positive or negative—for a given timestamp, but not both.

To encode boundary prompts, the binary value \( p_{b,t} \in \{0, 1\} \) is used to represent the boundary information at each timestamp, where \( p_{b,t} = 1 \) indicates the positive boundary (a state transition should occur) and \( p_{b,t} = 0 \) indicates the negative boundary (no state transition should occur).
This value is mapped into the embedding space of size \( D \):
\begin{equation}
\mathbf{e}_{b,t} = E_b(p_{b,t}),
\end{equation}
where \( E_b(\cdot) \) is a trainable lookup table, and \( \mathbf{e}_{b,t} \in \mathbb{R}^D\) denotes the embedding of the boundary prompt for timestamp \( t \).
To facilitate the direct combination with label prompt embeddings, we set the embedding size of boundary prompts to be the same as the embedding size of label prompts, both denoted as \( D \).
This ensures that label and boundary embeddings can be seamlessly added together in the next step.

\subsubsection{Combining Label and Boundary Prompt Embeddings}
\label{sec:combining_prompts}
To fully incorporate the information from both label prompts and boundary prompts, the embeddings \( \mathbf{e}_{l,t} \) and \( \mathbf{e}_{b,t} \) for each timestamp \( t \) are adjusted based on the availability of the corresponding prompts.
If a prompt is not provided, its embedding is effectively replaced with a zero vector.
The adjusted embeddings are then added together to form the final prompt embedding:
\begin{equation}
\mathbf{z}_{p,t} = \mathbb{I}(p_{l,t}) \mathbf{e}_{l,t} + \mathbb{I}(p_{b,t}) \mathbf{e}_{b,t}
\end{equation}
where \( \mathbb{I}(p_{l,t}) \) and \( \mathbb{I}(p_{b,t}) \) are indicator functions that take the value 1 if the respective prompt is provided, otherwise 0.
This ensures that only available prompts contribute to the final embedding while missing prompts are treated as zero vectors.
For timestamps where no prompts (neither label nor boundary) are provided, since both embeddings are set to a zero vector of dimension \( D \), their sum \( \mathbf{z}_{p,t} \) is also a zero vector of dimension \( D \).
This approach ensures that the model appropriately handles cases where prompts are not given.
The use of a zero embedding vector and an additive formulation simplifies the encoding process by eliminating the need for additional gating mechanisms or explicitly designated indicators, such as special tokens (e.g., [NA]), to signal the absence of a prompt.

\subsection{State Decoder}
\label{sec:state_decoder}
The goal of the state decoder (\( f_s \)) is to map the time series embeddings \( \mathbf{z}_x \) and prompt embeddings \( \mathbf{z}_p \) to the output states \( \mathbf{s} \).
To effectively leverage the information in these embeddings, we employ a two-way Transformer decoder, inspired by SAM \cite{sam} and other Transformer-based segmentation models \cite{transformer_seg_1,transformer_seg_2}.
Unlike conventional Transformer decoders that utilize a single direction of cross-attention, this design incorporates bidirectional interactions between the prompt and time series embeddings.

Given the time series embeddings \( \mathbf{z}_x^{(0)} \in \mathbb{R}^{T_{\text{patched}} \times D} \) and prompt embeddings \( \mathbf{z}_p^{(0)} \in \mathbb{R}^{T \times D} \), where the superscript \((0)\) indicates the initial representations before passing through any layers of the state decoder, the two-way Transformer decoder iteratively refines both embeddings over \( L_s \) stacked layers.
At each layer \( l \), the update process consists of the following steps:
(1) self-attention is applied to the prompt embeddings, enabling information exchange among prompts;
(2) cross-attention is performed with the prompt embeddings as queries and the time series embeddings as keys and values, allowing the prompts to extract relevant time series information;
(3) a bottleneck MLP refines the prompt embeddings, further processing their representation; and
(4) cross-attention is applied in the opposite direction, where the time series embeddings act as queries and the prompt embeddings serve as keys and values, incorporating prompt information into the time series embeddings.
For conciseness, we define \(\text{TWTL}\) (Two-Way Transformer Layer) as a single instance of this update process, which is applied iteratively across \( L_s \) layers.
The update at the layer \( l \) is expressed as:
\begin{equation}
\mathbf{z}_p^{(l)}, \mathbf{z}_x^{(l)} = \text{TWTL}(\mathbf{z}_p^{(l-1)}, \mathbf{z}_x^{(l-1)}), \quad l = 1, \dots, L_s.
\end{equation}
Finally, the updated prompt embeddings after \( L_s \) layers are passed through a linear layer \( \text{Linear}_s \), followed by a softmax function to obtain the final state probabilities:  

\begin{equation}
\mathbf{s} = \text{Softmax}(\text{Linear}_s(\mathbf{z}_p^{(L_s)})).
\end{equation}
Since \( \mathbf{z}_p^{(L_s)} \) integrates both time series patterns and prompt guidance, this formulation ensures that state predictions remain aligned with user-provided constraints while still being informed by the time series data.

\subsection{Iterative Training}
\label{sec:training}  
Since PromptTSS is designed as an interactive system where users provide prompts to control the desired output states, the training process must mimic this interactive behavior \cite{interactive_1, interactive_2}.
To achieve this, we adopt an iterative training strategy, allowing the model to progressively learn how to incorporate varying levels of user-provided prompts.  
Throughout the iterative training, all three core modules of PromptTSS—the time series encoder \( f_x \), the prompt encoder \( f_p \), and the state decoder \( f_s \)—are trained together.

At the beginning of each training iteration, no prompts are provided, meaning all prompt embeddings are replaced with zero vectors.
This setup forces the model to initially predict the output states \( \mathbf{s} = (s_1, \ldots, s_T) \) based solely on the time series embeddings \( \mathbf{z}_x \), ensuring that it does not become overly reliant on prompts.  

The training process then proceeds iteratively.
At each iteration \( r \), a new subset of prompts is sampled and added to the previously sampled prompts, reflecting the way users interact with the system by incrementally providing more guidance to refine the model’s predictions.
Specifically, the number of newly sampled prompts is drawn uniformly from a range \((n_{\text{min}}, n_{\text{max}})\), ensuring a gradual increase in available prompts.
Both label prompts \( \mathbf{p}_l^{(r)} \) and boundary prompts \( \mathbf{p}_b^{(r)} \) are sampled uniformly to maintain balanced exposure across training.  

Formally, at iteration \( r \), the prompt set is updated as:  
\begin{equation}
\mathbf{p}_l^{(r)} = \mathbf{p}_l^{(r-1)} \cup \Delta \mathbf{p}_l^{(r)}, \quad 
\mathbf{p}_b^{(r)} = \mathbf{p}_b^{(r-1)} \cup \Delta \mathbf{p}_b^{(r)},
\end{equation}  
where \( \Delta \mathbf{p}_l^{(r)} \) and \( \Delta \mathbf{p}_b^{(r)} \) represent the newly sampled label and boundary prompts, respectively.
The prompt embeddings \( \mathbf{z}_p \) are then updated accordingly.  
At each iteration, the model is optimized using cross-entropy loss \( \mathcal{L}_{CE} \) between the predicted states \( \hat{s}_t^{(r)} \) and the ground-truth states \( s_t \):  
\begin{equation}
\mathcal{L}_{\text{iter}}^{(r)} = \frac{1}{T} \sum_{t=1}^T \mathcal{L}_{CE}(s_t, \hat{s}_t^{(r)}),
\end{equation}  
The final training objective aggregates the loss across all iterations:  
\begin{equation}
\mathcal{L}_{\text{train}} = \sum_{r=1}^{N_r} \mathcal{L}_{\text{iter}}^{(r)},
\end{equation}  
where \( N_r \) denotes the total number of training iterations.  
% In our experiments, we set \( N_r = 8 \), as this was found to be sufficient for real-world use cases.
% Additional iterations were tested but did not yield significant performance gains.
This iterative approach allows the model to gradually learn how to effectively incorporate varying levels of user-provided prompts, enabling fine-grained control over state granularity and improving the model's adaptability to unseen patterns during inference.

\input{Tables/dataset}
\input{Tables/multiple_granularity}

\section{Experiments}
\label{sec:experiments}

\paragraph{Datasets}
We validate PromptTSS on six real-world datasets, with their statistics summarized in Table \ref{tab:dataset}.
Three of these datasets come from industrial manufacturing operations, while the other three cover diverse application scenarios such as human activity recognition and motion capture.
\textbf{Pump} \cite{prectime} contains multivariate time series from End-of-Line (EoL) testing of hydraulic pumps, with three subsets (V35, V36, V38) annotated by operational states.  
\textbf{MoCap}\footnote{\url{http://mocap.cs.cmu.edu/}} provides motion data from CMU's motion capture dataset; we use four dimensions for arms and legs following \cite{time2state}.  
\textbf{ActRecTut} \cite{act_rec_tut} includes acceleration data from sports activities, with our focus on walking-related subsets.  
\textbf{USC-HAD} \cite{usc_had} records human activities using a 3-axis accelerometer and gyrometer on the hip.
\textbf{PAMAP2} \cite{pamap2} contains human activity data from 8 subjects performing 24 low- and high-level actions.

To create datasets with multiple levels of state granularity, we systematically merge adjacent states in the original fine-grained states.
For Pump V35, Pump V36, and Pump V38, we generate both 2$\times$ and 4$\times$ coarser versions by progressively grouping consecutive states.
For USC-HAD, we only create a 2$\times$ coarser version because the original dataset contains only 12 states, and further merging would result in an excessively small number of states, limiting meaningful segmentation analysis.
We do not apply this state merging process to MoCap and ActRecTut, as their original number of states is already limited, and further merging would result in an insufficient number of states for robust analysis.

Unlike open datasets where coarse granularity is constructed synthetically, we also include \textbf{IndustryMG}, a proprietary industrial dataset with naturally annotated multi-granularity states (e.g., high-level production phases and fine-grained machine operations).
It serves as a real-world case study highlighting the practical relevance of multi-granularity segmentation.

\paragraph{Metrics}
To evaluate the segmentation performance of PromptTSS, we use three metrics: Accuracy (ACC), Macro F1-score (MF1), and Adjusted Rand Index (ARI).
These metrics provide a comprehensive assessment of the model's ability to correctly predict states, balance performance across classes, and capture clustering consistency.

\paragraph{Baselines}
We focus on a range of state-of-the-art models in time series segmentation.
\textbf{PrecTime} \cite{prectime} is a sequence-to-sequence model combining sliding windows and dense labeling for accurate industrial time series segmentation.  
\textbf{MS-TCN++} \cite{ms_tcn} is a multi-stage convolutional model using dilated convolutions to refine predictions and reduce over-segmentation.  
\textbf{U-Time} \cite{u_time} adapts a U-Net-based image segmentation architecture for time series segmentation.  
\textbf{DeepConvLSTM} \cite{deepconvlstm} combines CNNs for feature extraction with LSTMs for temporal modeling.

While time series forecasting models are not designed for segmentation, their strong temporal modeling makes them worth exploring.  
We adapt \textbf{iTransformer}~\cite{itransformer} and \textbf{PatchTST}~\cite{patchtst} by adding a projection layer for state probabilities and replacing the MSE loss with cross-entropy.  
These adapted models, denoted as iTransformer-TSS and PatchTST-TSS (where TSS stands for time series segmentation), are included in our comparison.

Unlike PromptTSS, baseline models do not inherently support multi-granularity prediction or prompt-guided inference.
PromptTSS is specifically designed to accept prompt signals during inference, using a dedicated prompt encoder and mask decoder to condition predictions on user guidance in real time.
In contrast, all baseline architectures lack modular components for conditioning on auxiliary inputs like prompts, making direct prompt integration non-trivial and beyond the scope of fair comparison.
To enable a comparable level of adaptability, we modify baselines using a post-hoc mechanism: for each model, we train separate sub-models at different levels of state granularity.
During inference, each sub-model produces a prediction, and we select the output that best matches the user-provided prompts.
This selection is based on how well predicted states align with provided label and boundary prompts.
While this allows baselines to respond to prompt constraints to some extent, it does not enable real-time prompt-aware refinement as in PromptTSS.

\begin{figure}[t]
    \centering
    \includegraphics[width=1.0\columnwidth]{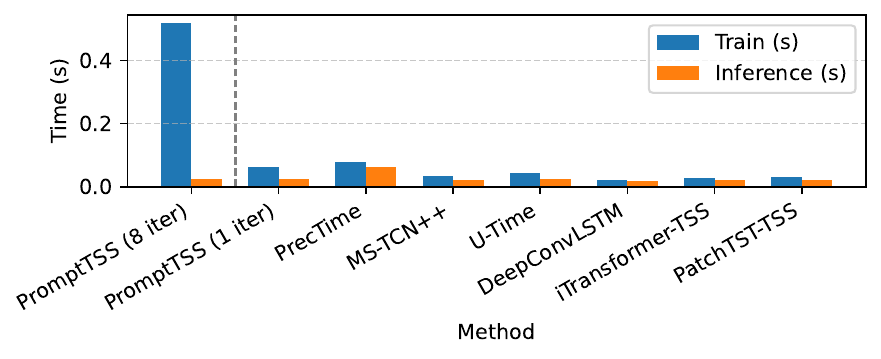}
    \caption{Training and inference time on USC-HAD dataset.}
    \label{fig:running_time}
\end{figure}

\paragraph{Implementation Details}  
We split each dataset chronologically into 70\% training, 15\% validation, and 15\% testing.  
This preserves temporal order and prevents data leakage, which could occur with random splitting in time series data.
For the sliding window length \(T\) and stride \(S\), we set \(T = 256\) for all datasets except USC-HAD, which requires a longer window (\(T = 512\)) due to its longer average state durations.  
Empirical findings from~\cite{time2state} suggest that setting \(S\) between 20\% and 40\% of \(T\) balances accuracy and efficiency.  
Following this, we use \(S = 64\) for datasets with \(T = 256\), and \(S = 128\) for USC-HAD.  
We confirmed that reducing \(S\) further did not yield accuracy improvements but substantially increased runtime.

For \emph{PromptTSS}, we use the AdamW optimizer with a weight decay of 0.01 and a fixed learning rate of 0.0001.  
The time series encoder has three Transformer layers, and the state decoder includes six two-way attention blocks, both with hidden size \(D = 128\).  
We apply a dropout rate of 0.1 and use early stopping based on validation performance.  
The patching mechanism uses a patch length of 16 and a stride of 8.  
During iterative training, the number of iterations is set to \(N_r = 8\), with \(n_{\min} = 1\) to \(n_{\max} = 3\) prompts randomly added in each iteration.

For \emph{baseline models}, we use official implementations when available.  
If no official code is provided, we reimplement the models based on the descriptions in their original papers and tune hyperparameters through random search within a reasonable space (e.g., learning rate, hidden size, dropout).  
All baseline models are trained using early stopping to ensure sufficient convergence.

\input{Tables/single_granularity}

\subsection{Multi-Granularity Segmentation}
For multi-granularity experiments, we evaluate on six datasets: USC-HAD and PAMAP2, each with two levels of granularity (original and a 2× coarser version obtained by merging adjacent states); Pump V35, V36, and V38, each with three levels (original, 2× coarser, and 4× coarser); and IndustryMG, which contains two naturally annotated levels of multi-granularity states without synthetic merging.
These datasets are selected because their original state definitions support meaningful coarsening or include inherent multi-granularity.
In contrast, MoCap and ActRecTut are excluded, as their limited number of states makes further granularity reduction impractical.
In all experiments, prompts are provided for 5\% of the time steps in each sliding window (\(T\)), using both label and boundary prompts.
For instance, with \(T = 256\), prompts are given for 12 time steps to guide the segmentation process.
This setting allows us to assess how well PromptTSS generalizes and infers hierarchical state transitions with limited supervision.

In Table \ref{tab:multiple_granularity}, PromptTSS significantly outperforms all baselines, achieving an average accuracy improvement of 24.49\% compared to the best-performing baseline for each dataset.
This demonstrates its superior adaptability in segmenting time series across multiple granularity levels.
While baselines struggle with state transitions, PromptTSS dynamically adjusts using prompt-based guidance to maintain segmentation accuracy. 
Although baselines are adapted to support multi-granularity and prompt-guided inference, their prompts serve only as a post-hoc selection mechanism rather than influencing the segmentation process directly. 
This limits their ability to refine predictions based on user guidance. 
In contrast, PromptTSS integrates prompts as a direct input, enabling real-time adaptation and significantly improving segmentation performance.

% Among baselines, PrecTime performs poorly on multi-granularity datasets like Pump V35, V36, and V38, despite being designed for industrial data. It even trails U-Time and DeepConvLSTM, suggesting limitations in handling hierarchical state structures.  
% Among forecasting variants, iTransformer-TSS performs the worst, even behind PatchTST-TSS, despite its stronger forecasting capabilities.  
% This highlights a key limitation: models optimized for continuous prediction do not generalize well to discrete segmentation, especially in multi-granularity settings where hierarchical transitions add complexity.

Among baselines, PrecTime underperforms on multi-granularity datasets like Pump V35, V36, and V38, despite being designed for industrial data, and performs worse than U-Time and DeepConvLSTM.  
This suggests limitations in its ability to model multi-level state transitions.  
iTransformer-TSS performs the worst overall, even though it is more advanced in forecasting, highlighting that models built for continuous prediction do not transfer well to discrete segmentation, especially in multi-granularity settings.

% Despite using a Transformer-based encoder, PromptTSS maintains competitive efficiency during both training and inference on the USC-HAD dataset.
% As shown in Figure~\ref{fig:running_time}, its patching mechanism reduces time complexity quadratically with respect to input length, enabling efficient computation—particularly in the 1-iteration setting.
% For most experiments, we use 8 training iterations to simulate realistic interactive scenarios.
% This design choice affects training time but does not impact inference speed.

PromptTSS maintains competitive training and inference efficiency despite using a Transformer-based encoder.  
As shown in Figure~\ref{fig:running_time}, its patching mechanism reduces time complexity quadratically with respect to input length, enabling fast computation, particularly in the 1-iteration setting.  
We use 8 training iterations in most experiments to simulate realistic interactive scenarios.  
While this iterative setup increases training time, it does not affect inference speed, which remains fast and consistent across settings.

\input{Tables/transfer_learning}

\subsection{Single-Granularity Segmentation}
For single-granularity experiments, we use all eight datasets with their original granularity. 
Similar to the multi-granularity setting, PromptTSS has access to prompts for 5\% of the total time steps in each sliding window. 
However, since baseline models rely on a post-hoc prompting mechanism, they cannot leverage prompts effectively in this setting. 
With only one level of states, there is no need to select a corresponding submodel, effectively rendering prompts unusable for the baselines.  

In Table \ref{tab:single_granularity}, PromptTSS outperforms all baselines, achieving an average accuracy improvement of 17.88\% compared to the best-performing baseline for each dataset. 
While its dominance remains clear, the performance gap is less pronounced compared to multi-granularity segmentation. 
This is expected, as PromptTSS excels not only due to its natural integration of prompting but also because of its inherent ability to handle multiple levels of state granularity.

Among the baselines, U-Time remains the second-best model overall, as observed in the multi-granularity experiments, though DeepConvLSTM performs best on ActRecTut, likely due to the dataset's small size of only two time series, which reduces variability and makes it less challenging to model.
Meanwhile, iTransformer-TSS continues to underperform, ranking below PatchTST-TSS, mirroring its struggles in the multi-granularity setting.

\begin{figure}[t]
    \centering
    \includegraphics[width=1.0\columnwidth]{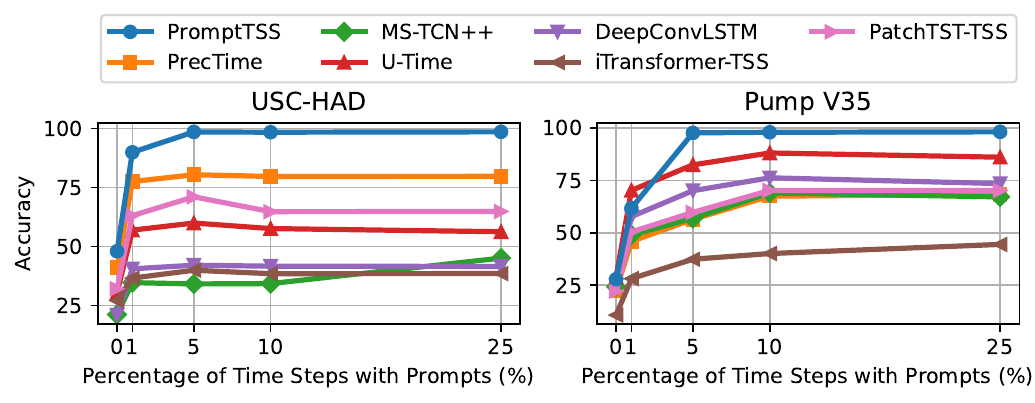}
    \caption{Ablation Study: Percentage of Time Steps with Prompts During Inference.}
    \label{fig:ablation_1}
\end{figure}

\subsection{Transfer Learning}

\subsubsection{Transfer Learning on Unseen Datasets}
To assess the robustness of PromptTSS, we test its ability to generalize to related but unseen datasets without retraining.  
We use the Pump dataset, which includes three subsets—V35, V36, and V38—capturing different pump types during End-of-Line (EoL) testing.  
While structurally similar, these subsets differ in sensor patterns and state transitions due to varying operating conditions.  
We train PromptTSS on Pump V35 and test it on V36 and V38 using a multi-granularity setup, with prompts provided for 5\% of time steps per sliding window.

As shown in Table~\ref{tab:transfer_learning} (top half), all baselines perform poorly in this setting, with the second-best model reaching only 22.19\% accuracy.  
In contrast, PromptTSS achieves 75.31\%, a 239.36\% improvement.  
This demonstrates PromptTSS’s ability to adapt to new datasets using minimal supervision, reducing annotation costs through high-quality pseudo-labeling guided by prompts.

\subsubsection{Transfer Learning on Unseen Granularity Levels}
We also evaluate whether PromptTSS can transfer knowledge across unseen granularity levels—a practical need in real-world annotation, where labeling at every level is often infeasible.  
Using Pump V35, we train on certain granularity levels and test on others, again using 5\% prompts to guide inference.  
For example, one scenario involves training on original and 2× coarser states and testing on 4× coarser.

Table~\ref{tab:transfer_learning} (bottom half) shows that baselines struggle to generalize, with the second-best model achieving just 8.81\% accuracy.  
PromptTSS achieves 61.60\%, a 599.24\% improvement.  
Though the absolute accuracy is lower than in the unseen dataset setting, the ability to generalize across granularity levels significantly reduces labeling effort.  
In a harder case—training on coarse (2× and 4×) granularities and testing on fine-grained states—PromptTSS achieves 38.59\% accuracy, as expected due to the difficulty of recovering detailed states from abstract ones.

We acknowledge the improvements may appear surprising, but they result from two core innovations in PromptTSS: a unified model trained across granularities and a prompt-based mechanism for real-time adaptation.  
In contrast, baselines rely on static sub-models without dynamic guidance, limiting their generalization in unseen scenarios.

\begin{figure}[t]
    \centering
    \includegraphics[width=1.0\columnwidth]{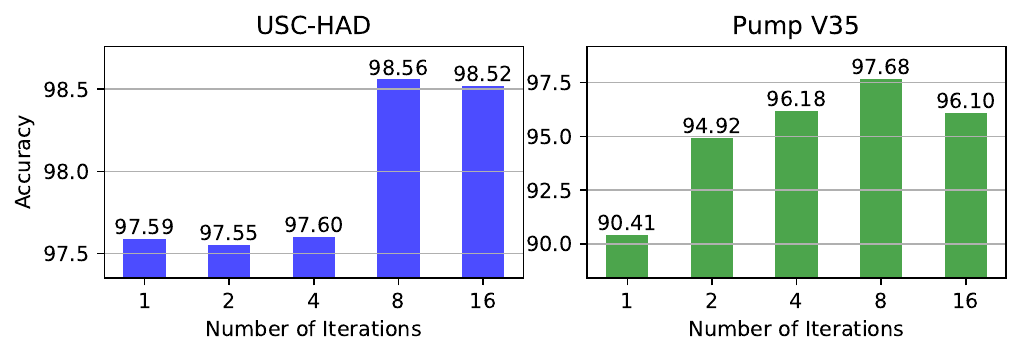}
    \caption{Ablation Study: Number of Iterations in Iterative Training.}
    \label{fig:ablation_2}
\end{figure}

\subsection{Ablation Study}
\subsubsection{Percentage of Time Steps with Prompts During Inference}

To assess the adaptability of PromptTSS under different levels of supervision, we evaluate its performance across varying percentages of time steps with prompts, as shown in Figure~\ref{fig:ablation_1}.  
At 0\% prompts, all models suffer from poor segmentation performance due to the lack of guidance—especially in multi-granularity settings where multiple plausible state assignments exist.  
Despite this, PromptTSS still outperforms all baselines, achieving a 66.73\% accuracy improvement on USC-HAD and 25.62\% on Pump V35.  
This strong performance is attributed to its unified multi-granularity training, which allows the model to capture both coarse and fine temporal patterns.  
In contrast, baseline models rely on separately trained sub-models and fail to generalize across different granularities.

% However, even with a small fraction of prompts, PromptTSS shows a strong ability to leverage minimal supervision for accurate segmentation.
% With just 1\% of time steps provided with prompts, the model achieves 89.93\% accuracy on the USC-HAD dataset.
% Given the experimental setting of \(T=512\), this corresponds to just five prompted time steps per window—demonstrating the efficiency of interactive guidance.
% As the percentage of prompts increases, performance improves rapidly.
% With 5\% prompts, PromptTSS exceeds 95\% accuracy, significantly reducing the need for additional supervision.
% Increasing to 10\% or 25\% yields only marginal gains, making higher prompt densities impractical for real-time use.
% This saturation effect, also observed in baselines, suggests that once sufficient guidance is available, further prompting offers diminishing returns.
% PromptTSS is designed to achieve near-optimal accuracy with minimal user input, making it highly suitable for interactive settings.
\sloppy
While performance degrades for all models at 0\% prompts, PromptTSS quickly recovers with minimal supervision.
Even at just 1\% of time steps prompted, it achieves 89.93\% accuracy on the USC-HAD dataset.  
Given \(T=512\), this translates to only five prompted time steps per window—highlighting the efficiency of interactive guidance.  
As the prompt percentage increases, performance improves rapidly: at 5\%, PromptTSS exceeds 95\% accuracy, substantially reducing the need for further supervision.  
Beyond this point, increasing to 10\% or 25\% yields only marginal gains, making higher prompt densities impractical for real-time use.  
This saturation effect, also observed in baselines, suggests that once sufficient guidance is provided, additional prompts bring diminishing returns.  
PromptTSS is intentionally designed to reach near-optimal accuracy with minimal user input, making it highly suitable for interactive segmentation tasks.

While baseline models are adapted to support prompt-guided inference, their use of prompts differs fundamentally from PromptTSS.
PromptTSS incorporates prompts directly into its segmentation pipeline through a prompt encoder and a mask decoder, enabling real-time refinement.
In contrast, baselines lack native support for prompts and instead rely on post-hoc model selection, choosing among pre-trained submodels based on prompt consistency.
This structural difference explains why PromptTSS shows a sharp performance gain even with only 5\% prompts in Figure~\ref{fig:ablation_1}.

\begin{figure}[t]
    \centering
    \includegraphics[width=1.0\columnwidth]{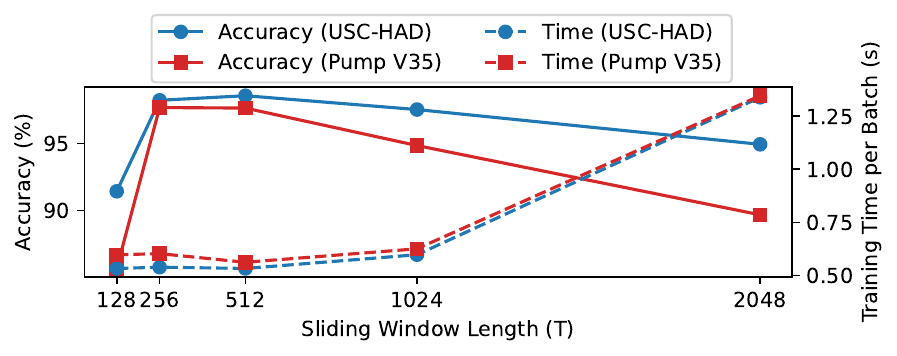}
    \caption{Ablation Study: Effect of sliding window length \(T\) on accuracy and training time for USC-HAD and PumpV35.}
    \label{fig:sliding_window_length}
\end{figure}

\subsubsection{Number of Iterations in Iterative Training}
To evaluate the impact of iterative training on segmentation performance, we analyze how the number of iterations affects PromptTSS.
Since PromptTSS is designed as an interactive segmentation tool, its training process mimics user behavior, where prompts are not provided all at once but instead in multiple steps.
This iterative training strategy allows the model to progressively refine its segmentation based on incremental supervision.

As shown in Figure \ref{fig:ablation_2}, there is a clear trend where increasing the number of training iterations leads to higher accuracy.
This improvement occurs because each iteration introduces new prompts, helping the model learn to adjust its segmentation outputs dynamically, just as a real user would provide guidance step by step.
However, performance does not improve indefinitely.
We observe that accuracy peaks at 8 iterations and then begins to decline at 16 iterations.
This decline occurs because a higher number of iterations results in more provided prompts, which can cause the model to over-rely on user guidance rather than learning robust segmentation patterns.
Additionally, training with 16 iterations is computationally expensive, making it impractical for real-world applications.
Overall, these findings highlight that while iterative training enhances performance, excessive iterations can diminish generalization and increase computational costs.
% Based on this, we identify 8 iterations as the optimal balance between segmentation accuracy and efficiency.

\subsubsection{Sliding Window Length}
The effect of sliding window length \(T\) is evaluated in terms of segmentation accuracy and training efficiency.
As shown in Figure~\ref{fig:sliding_window_length}, extremely short windows (e.g., \(T = 128\)) limit temporal context, leading to lower segmentation accuracy.
Accuracy improves with larger window sizes as the model captures richer temporal information.
PromptTSS maintains high accuracy even with large \(T\), achieving strong performance at \(T = 2048\) while incurring minimal additional training time.
Thanks to its patching mechanism, training remains efficient across all window lengths, demonstrating PromptTSS scales well to long input sequences.

\begin{figure}[t]
    \centering
    \includegraphics[width=1.0\columnwidth]{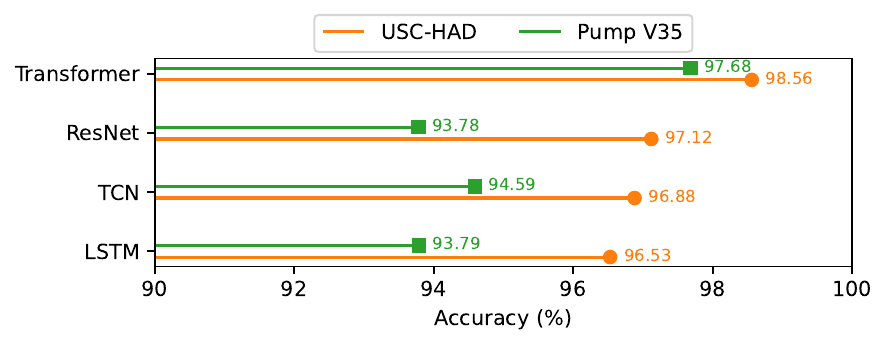}
    \caption{Ablation Study: Comparison of different encoder architectures on USC-HAD and PumpV35.}
    \label{fig:encoder_arch}
\end{figure}

% \subsubsection{Encoder Architecture}
% PromptTSS is evaluated with different backbone encoders: Transformer, ResNet, TCN, and LSTM.
% As shown in Figure~\ref{fig:encoder_arch}, the Transformer consistently outperforms the other architectures on both USC-HAD and PumpV35.
% This result highlights its superior ability to model long-range temporal dependencies, justifying its selection as the default encoder for PromptTSS.

\subsubsection{Encoder Architecture}
PromptTSS is evaluated with different backbone encoders: Transformer, ResNet, TCN, and LSTM. 
As shown in Figure~\ref{fig:encoder_arch}, the Transformer consistently outperforms others on USC-HAD and PumpV35, demonstrating superior long-range temporal modeling and justifying its use as the default encoder.

\section{Conclusion}
\label{sec:conclusion}
In this work, we introduce PromptTSS, a novel prompting-based framework for interactive multi-granularity time series segmentation.
PromptTSS addresses two major challenges in time series segmentation: (1) handling multiple levels of granularity within a unified model and (2) adapting dynamically to new and evolving patterns in real-world applications.
Our approach leverages a prompting mechanism that incorporates label and boundary information to guide segmentation.
This allows PromptTSS to seamlessly integrate hierarchical state structures while dynamically refining segmentation outputs based on user-provided prompts.
Through extensive experiments, we demonstrated significant performance gains, achieving 24.49\% and 17.88\% accuracy improvements in multi- and single-granularity segmentation, respectively.
Additionally, PromptTSS exhibited strong transfer learning capabilities, generalizing to unseen datasets and state granularities, with up to a 599.24\% accuracy improvement in transfer scenarios.
Our ablation studies further confirmed its efficiency, requiring as little as 1\% of time steps with prompts to achieve high accuracy.
With its iterative training strategy, PromptTSS progressively refines predictions, making it a robust, adaptive, and practical solution for real-world applications.

In future work, we will explore transitioning PromptTSS from a per-timestep segmentation approach to a more structured, mask-based segmentation framework.
Additionally, PromptTSS could benefit from incorporating temporal consistency mechanisms, allowing prompts to persist and influence across the sliding window rather than being applied independently.

\begin{acks}
The authors would like to thank GoEdge.ai for providing internship support, computing resources, and valuable domain expertise that contributed to this research.
\end{acks}
% \begin{acks}
% The authors thank GoEdge.ai for internship support, computing resources, and domain expertise contributing to this research.
% \end{acks}

% TODO: remove for arxiv
\section*{GenAI Usage Disclosure}
In accordance with the ACM policy and CIKM 2025 submission guidelines, we disclose the following use of Generative AI in the preparation of this work: Generative AI tools (e.g., ChatGPT) were used solely for language polishing purposes, including minor grammar correction, sentence restructuring, and clarity improvements. No content, code, or experimental results were generated or influenced by these tools.

\bibliographystyle{ACM-Reference-Format}
\bibliography{reference}

\end{document}

%% file: Tables/dataset.tex
\renewcommand{\arraystretch}{1}  % Row spacing (default: 1)
\setlength{\tabcolsep}{2pt}       % Col spacing (default: 6pt)

\begin{table}[t]
% \tiny
\centering
\caption{Statistical overview of the 8 datasets for time-series segmentation.}
\label{tab:dataset}
\resizebox{\linewidth}{!}{
\begin{tabular}{c|cccccc}
\hline
Datasets & \# Features & \# Timesteps & \multicolumn{1}{c}{\# Time Series} & \# States & State Duration & Avg State Duration \\ \hline
Pump V35 & 9 & 27,770 $\sim$ 40,810 & \multicolumn{1}{c}{40} & 41 $\sim$ 43 & 1 $\sim$ 3,290 & 543 \\
Pump V36 & 9 & 26,185 $\sim$ 38,027 & \multicolumn{1}{c}{40} & 41 $\sim$ 43 & 1 $\sim$ 1,960 & 517 \\
Pump V38 & 9 & 20,365 $\sim$ 30,300 & \multicolumn{1}{c}{40} & 42 $\sim$ 43 & 1 $\sim$ 1,820 & 407 \\
MoCap & 4 & 4,579 $\sim$ 10,617 & \multicolumn{1}{c}{9} & 3 $\sim$ 9 & 354 $\sim$ 2,003 & 886 \\
ActRecTut & 23 & 31,392 $\sim$ 32,577 & \multicolumn{1}{c}{2} & 6 & 22 $\sim$ 5,100 & 743 \\
USC-HAD & 6 & 25,356 $\sim$ 56,251 & \multicolumn{1}{c}{70} & 12 & 600 $\sim$ 13,500 & 3,347 \\
PAMAP2 & 9 & 8,477 $\sim$ 447,000 & \multicolumn{1}{c}{9} & 2 $\sim$ 13 & 1 $\sim$ 42,995 & 14,434 \\
IndustryMG (Fine-Grained) & 3 & 13,629 $\sim$ 45,010 & \multicolumn{1}{c}{17} & 4 & 10 $\sim$ 2,236 & 583 \\ \hline
Pump V35 (2x Coarser) & \multicolumn{1}{r}{----------} & Same as PumpV35 & \multicolumn{1}{l}{----------} & 22 & 1 $\sim$ 3,570 & 771 \\
Pump V36 (2x Coarser) & \multicolumn{1}{r}{----------} & Same as PumpV36 & \multicolumn{1}{l}{----------} & 21 $\sim$ 22 & 1 $\sim$ 2,085 & 719 \\
Pump V38 (2x Coarser) & \multicolumn{1}{r}{----------} & Same as PumpV38 & \multicolumn{1}{l}{----------} & 22 & 1 $\sim$ 3,305 & 625 \\
USC-HAD (2x Coarser) & \multicolumn{1}{r}{----------} & Same as USC-HAD & \multicolumn{1}{l}{----------} & 6 & 1,700 $\sim$ 19,100 & 6,694 \\
PAMAP2 (2x Coarser) & \multicolumn{1}{r}{----------} & Same as PAMAP2 & \multicolumn{1}{l}{----------} & 6 & 1 $\sim$ 50,198 & 16,997 \\
IndustryMG (Coarse-Grained) & \multicolumn{1}{r}{----------} & \begin{tabular}[c]{@{}c@{}}Same as IndustryMG\\ (Fine-Grained)\end{tabular} & \multicolumn{1}{l}{----------} & 2 & 18 $\sim$ 3,415 & 883 \\ \hline
Pump V35 (4x Coarser) & \multicolumn{1}{r}{----------} & Same as PumpV35 & \multicolumn{1}{l}{----------} & 11 & 1 $\sim$ 3,855 & 1,171 \\
Pump V36 (4x Coarser) & \multicolumn{1}{r}{----------} & Same as PumpV36 & \multicolumn{1}{l}{----------} & 11 & 1 $\sim$ 4,874 & 1,019 \\
Pump V38 (4x Coarser) & \multicolumn{1}{r}{----------} & Same as PumpV38 & \multicolumn{1}{l}{----------} & 11 & 1 $\sim$ 4,154 & 933 \\ \hline
\end{tabular}
}
\end{table}

%% file: Tables/multiple_granularity.tex
\renewcommand{\arraystretch}{0.85}  % Row spacing (default: 1)
\setlength{\tabcolsep}{6pt}       % Col spacing (default: 6pt)

\begin{table*}
\tiny
\centering
\caption{Segmentation performance on multi-granularity datasets. Best results are in {\color[HTML]{FF0000} \textbf{bold}}, and second-best are {\color[HTML]{0000FF} {\ul underlined}}.}
\label{tab:multiple_granularity}
\resizebox{\linewidth}{!}{
\begin{tabular}{c|c|c|c|c|c|c|c|c}
\hline
Dataset & Metric & PromptTSS & PrecTime & MS-TCN++ & U-Time & DeepConvLSTM & iTransformer-TSS & PatchTST-TSS \\ \hline
 & ACC & {\color[HTML]{FF0000} \textbf{98.56}} & {\color[HTML]{0000FF} {\ul 82.82}} & 47.43 & 67.39 & 54.03 & 45.86 & 77.72 \\
 & MF1 & {\color[HTML]{FF0000} \textbf{98.00}} & {\color[HTML]{0000FF} {\ul 75.68}} & 43.57 & 58.84 & 47.43 & 31.95 & 63.17 \\
\multirow{-3}{*}{\begin{tabular}[c]{@{}c@{}}USC-HAD\\ (Original + 2x Coarser)\end{tabular}} & ARI & {\color[HTML]{FF0000} \textbf{97.19}} & {\color[HTML]{0000FF} {\ul 68.52}} & 25.82 & 48.40 & 31.18 & 24.53 & 63.71 \\ \hline

 & ACC & {\color[HTML]{FF0000} \textbf{97.68}} & 56.37 & 56.88 & {\color[HTML]{0000FF} {\ul 82.41}} & 70.09 & 37.36 & 59.72 \\
 & MF1 & {\color[HTML]{FF0000} \textbf{95.53}} & 54.10 & 42.30 & {\color[HTML]{0000FF} {\ul 78.75}} & 62.58 & 20.91 & 48.70 \\
\multirow{-3}{*}{\begin{tabular}[c]{@{}c@{}}Pump V35\\ (Original + 2x + 4x Coarser)\end{tabular}} & ARI & {\color[HTML]{FF0000} \textbf{95.98}} & 35.63 & 41.59 & {\color[HTML]{0000FF} {\ul 71.89}} & 56.31 & 19.72 & 43.11 \\ \hline

 & ACC & {\color[HTML]{FF0000} \textbf{96.48}} & 62.82 & 48.85 & {\color[HTML]{0000FF} {\ul 80.04}} & 63.63 & 37.60 & 57.46 \\
 & MF1 & {\color[HTML]{FF0000} \textbf{93.77}} & 54.53 & 39.17 & {\color[HTML]{0000FF} {\ul 71.30}} & 59.14 & 23.40 & 44.65 \\
\multirow{-3}{*}{\begin{tabular}[c]{@{}c@{}}Pump V36\\ (Original + 2x + 4x Coarser)\end{tabular}} & ARI & {\color[HTML]{FF0000} \textbf{93.80}} & 41.60 & 28.81 & {\color[HTML]{0000FF} {\ul 69.62}} & 42.94 & 16.91 & 36.63 \\ \hline

 & ACC & {\color[HTML]{FF0000} \textbf{96.23}} & 62.37 & 45.40 & {\color[HTML]{0000FF} {\ul 75.91}} & 65.67 & 34.27 & 57.16 \\
 & MF1 & {\color[HTML]{FF0000} \textbf{93.56}} & 60.44 & 25.91 & {\color[HTML]{0000FF} {\ul 69.79}} & 61.10 & 17.47 & 44.36 \\
\multirow{-3}{*}{\begin{tabular}[c]{@{}c@{}}Pump V38\\ (Original + 2x + 4x Coarser)\end{tabular}} & ARI & {\color[HTML]{FF0000} \textbf{92.75}} & 43.13 & 28.72 & {\color[HTML]{0000FF} {\ul 62.06}} & 45.13 & 16.94 & 36.17 \\ \hline

 & ACC & {\color[HTML]{FF0000} \textbf{99.05}} & 52.47 & 48.54 & 54.72 & {\color[HTML]{0000FF} {\ul 61.31}} & 41.39 & 42.76 \\
 & MF1 & {\color[HTML]{FF0000} \textbf{98.78}} & 40.43 & 16.89 & 41.03 & {\color[HTML]{0000FF} {\ul 50.67}} & 11.26 & 27.96 \\
\multirow{-3}{*}{\begin{tabular}[c]{@{}c@{}}PAMAP2\\ (Original + 2x Coarser)\end{tabular}} & ARI & {\color[HTML]{FF0000} \textbf{97.46}} & 19.71 & 16.54 & 24.38 & {\color[HTML]{0000FF} {\ul 35.60}} & 9.56 & 9.93 \\ \hline

 & ACC & {\color[HTML]{FF0000} \textbf{96.35}} & 82.74 & {\color[HTML]{0000FF} {\ul 86.91}} & 70.52 & 83.53 & 68.69 & 27.48 \\
 & MF1 & {\color[HTML]{FF0000} \textbf{77.10}} & {\color[HTML]{0000FF} {\ul 64.59}} & 48.58 & 47.92 & 64.35 & 24.80 & 24.82 \\
\multirow{-3}{*}{\begin{tabular}[c]{@{}c@{}}IndustryMG\\ (Fine + Coarse Levels)\end{tabular}} & ARI & {\color[HTML]{FF0000} \textbf{90.93}} & 52.34 & {\color[HTML]{0000FF} {\ul 60.77}} & 33.05 & 54.19 & 24.54 & 22.11 \\ \hline
\end{tabular}
}
\end{table*}

%% file: Tables/single_granularity.tex
\renewcommand{\arraystretch}{0.85}  % Row spacing (default: 1)
\setlength{\tabcolsep}{10pt}       % Col spacing (default: 6pt)

\begin{table*}
\tiny
\centering
\caption{Segmentation performance on single-granularity datasets. Best results are in {\color[HTML]{FF0000} \textbf{bold}}, and second-best are {\color[HTML]{0000FF} {\ul underlined}}.}
\label{tab:single_granularity}
\resizebox{\linewidth}{!}{
\begin{tabular}{c|c|c|c|c|c|c|c|c}
\hline
Dataset & Metric & PromptTSS & PrecTime & MS-TCN++ & U-Time & DeepConvLSTM & iTransformer-TSS & PatchTST-TSS \\ \hline
 & ACC & {\color[HTML]{FF0000} \textbf{91.86}} & 39.77 & 58.40 & 52.78 & {\color[HTML]{0000FF} {\ul 61.35}} & 45.84 & 52.99 \\
 & MF1 & {\color[HTML]{FF0000} \textbf{91.31}} & 19.37 & 28.29 & 18.76 & {\color[HTML]{0000FF} {\ul 35.34}} & 19.67 & 20.05 \\
\multirow{-3}{*}{MoCap} & ARI & {\color[HTML]{FF0000} \textbf{78.94}} & 3.12 & 27.88 & 21.71 & {\color[HTML]{0000FF} {\ul 32.72}} & 12.32 & 26.39 \\ \hline

 & ACC & 96.21 & 91.59 & 95.87 & {\color[HTML]{0000FF} {\ul 97.28}} & {\color[HTML]{FF0000} \textbf{98.14}} & 79.88 & 81.38 \\
 & MF1 & {\color[HTML]{0000FF} {\ul 84.04}} & 71.69 & 71.66 & 69.53 & {\color[HTML]{FF0000} \textbf{90.61}} & 60.51 & 61.34 \\
\multirow{-3}{*}{ActRecTut} & ARI & 93.60 & 79.37 & 94.25 & {\color[HTML]{0000FF} {\ul 96.41}} & {\color[HTML]{FF0000} \textbf{96.76}} & 51.49 & 54.71 \\ \hline

 & ACC & {\color[HTML]{FF0000} \textbf{96.56}} & {\color[HTML]{0000FF} {\ul 75.69}} & 37.95 & 63.32 & 49.89 & 42.68 & 70.35 \\
 & MF1 & {\color[HTML]{FF0000} \textbf{95.97}} & {\color[HTML]{0000FF} {\ul 71.98}} & 29.74 & 59.00 & 50.26 & 27.07 & 63.34 \\
\multirow{-3}{*}{USC-HAD} & ARI & {\color[HTML]{FF0000} \textbf{93.45}} & {\color[HTML]{0000FF} {\ul 62.77}} & 20.36 & 46.34 & 32.28 & 32.88 & 53.37 \\ \hline

 & ACC & {\color[HTML]{FF0000} \textbf{98.29}} & 72.51 & 69.83 & {\color[HTML]{0000FF} {\ul 91.86}} & 85.60 & 55.30 & 73.04 \\
 & MF1 & {\color[HTML]{FF0000} \textbf{94.28}} & 71.96 & 50.05 & {\color[HTML]{0000FF} {\ul 86.93}} & 80.27 & 35.14 & 54.79 \\
\multirow{-3}{*}{Pump V35} & ARI & {\color[HTML]{FF0000} \textbf{97.82}} & 70.07 & 70.49 & {\color[HTML]{0000FF} {\ul 89.59}} & 82.43 & 46.49 & 70.92 \\ \hline

 & ACC & {\color[HTML]{FF0000} \textbf{97.18}} & 77.38 & 67.81 & {\color[HTML]{0000FF} {\ul 87.20}} & 80.97 & 61.38 & 80.14 \\
 & MF1 & {\color[HTML]{FF0000} \textbf{93.69}} & 75.43 & 45.63 & {\color[HTML]{0000FF} {\ul 83.45}} & 76.47 & 43.40 & 61.00 \\
\multirow{-3}{*}{Pump V36} & ARI & {\color[HTML]{FF0000} \textbf{96.16}} & 68.96 & 62.07 & {\color[HTML]{0000FF} {\ul 81.47}} & 74.57 & 49.77 & 75.94 \\ \hline

 & ACC & {\color[HTML]{FF0000} \textbf{95.49}} & 75.19 & 55.88 & {\color[HTML]{0000FF} {\ul 84.82}} & 77.37 & 62.64 & 77.19 \\
 & MF1 & {\color[HTML]{FF0000} \textbf{91.08}} & 72.39 & 34.47 & {\color[HTML]{0000FF} {\ul 79.22}} & 71.72 & 43.88 & 68.19 \\
\multirow{-3}{*}{Pump V38} & ARI & {\color[HTML]{FF0000} \textbf{92.05}} & 67.55 & 53.81 & {\color[HTML]{0000FF} {\ul 80.78}} & 72.48 & 54.75 & 70.51 \\ \hline

 & ACC & {\color[HTML]{FF0000} \textbf{99.00}} & 50.40 & 41.97 & 53.91 & {\color[HTML]{0000FF} {\ul 58.52}} & 37.87 & 40.82 \\
 & MF1 & {\color[HTML]{FF0000} \textbf{98.91}} & 34.09 & 12.87 & 43.58 & {\color[HTML]{0000FF} {\ul 45.00}} & 17.01 & 24.27 \\
\multirow{-3}{*}{PAMAP2} & ARI & {\color[HTML]{FF0000} \textbf{97.31}} & 17.34 & 9.24 & 25.92 & {\color[HTML]{0000FF} {\ul 34.40}} & 12.34 & 14.02 \\ \hline

 & ACC &  {\color[HTML]{FF0000} \textbf{97.79}} & 95.92 & 91.99 & 95.46 & {\color[HTML]{0000FF} {\ul 97.65}} & 51.10 & 57.89 \\
 & MF1 & {\color[HTML]{FF0000} \textbf{89.91}} & 72.16 & 56.30 & 79.17 & {\color[HTML]{0000FF} {\ul 89.35}} & 19.41 & 41.10 \\
\multirow{-3}{*}{\begin{tabular}[c]{@{}c@{}}IndustryMG\\ (Fine-Grained Only)\end{tabular}} & ARI & {\color[HTML]{0000FF} {\ul 95.55}} & 91.27 & 84.51 & 91.79 & {\color[HTML]{FF0000} \textbf{95.62}} & 11.96 & 57.60 \\ \hline
\end{tabular}
}
\end{table*}

%% file: Tables/transfer_learning.tex
\renewcommand{\arraystretch}{0.85}  % Row spacing (default: 1)
\setlength{\tabcolsep}{10pt}       % Col spacing (default: 6pt)

\begin{table*}
\small
\centering
\caption{Segmentation performance in transfer learning. The top half reports results on unseen datasets, while the bottom half presents results on unseen granularity levels within the same dataset. Best results are in {\color[HTML]{FF0000} \textbf{bold}}, and second-best are {\color[HTML]{0000FF} {\ul underlined}}.}
\label{tab:transfer_learning}
\resizebox{\linewidth}{!}{
\begin{tabular}{c|c|c|c|c|c|c|c|c|c}
\hline
Setting & Dataset & Metric & PromptTSS & PrecTime & MS-TCN++ & U-Time & DeepConvLSTM & iTransformer-TSS & PatchTST-TSS \\ \hline
 &  & ACC & {\color[HTML]{FF0000} \textbf{73.91}} & 13.32 & 14.37 & 14.45 & {\color[HTML]{0000FF} {\ul 15.51}} & 11.99 & 12.86 \\
 &  & MF1 & {\color[HTML]{FF0000} \textbf{52.69}} & 5.74 & 5.62 & 6.47 & {\color[HTML]{0000FF} {\ul 6.50}} & 3.02 & 5.44 \\
 & \multirow{-3}{*}{\begin{tabular}[c]{@{}c@{}}Pump V35 (Original, 2×, 4× Coarser)\\      ↓\\      Pump V36 (Original, 2×, 4× Coarser)\end{tabular}} & ARI & {\color[HTML]{FF0000} \textbf{65.78}} & 10.72 & 9.87 & {\color[HTML]{0000FF} {\ul 11.52}} & 10.11 & 3.48 & 9.80 \\ \cline{2-10} 
 &  & ACC & {\color[HTML]{FF0000} \textbf{76.70}} & 21.16 & 21.57 & {\color[HTML]{0000FF} {\ul 28.87}} & 24.97 & 15.23 & 25.75 \\
 &  & MF1 & {\color[HTML]{FF0000} \textbf{56.82}} & 13.69 & 8.87 & {\color[HTML]{0000FF} {\ul 16.23}} & 13.02 & 6.26 & 16.02 \\
\multirow{-6}{*}{\begin{tabular}[c]{@{}c@{}}Unseen\\Datasets\end{tabular}} & \multirow{-3}{*}{\begin{tabular}[c]{@{}c@{}}Pump V35 (Original, 2×, 4× Coarser)\\      ↓\\      Pump V38 (Original, 2×, 4× Coarser)\end{tabular}} & ARI & {\color[HTML]{FF0000} \textbf{66.57}} & 13.32 & 19.50 & {\color[HTML]{0000FF} {\ul 21.89}} & 21.85 & 6.98 & 15.02 \\
\hhline{=|=|=|=|=|=|=|=|=|=}
 &  & ACC & {\color[HTML]{FF0000} \textbf{78.06}} & 5.01 & 7.14 & 4.11 & 7.82 & {\color[HTML]{0000FF} {\ul 9.68}} & 6.85 \\
 &  & MF1 & {\color[HTML]{FF0000} \textbf{23.87}} & 1.09 & {\color[HTML]{0000FF} {\ul 2.41}} & 1.54 & 2.18 & 1.67 & 2.22 \\
 & \multirow{-3}{*}{\begin{tabular}[c]{@{}c@{}}Pump V35 (Original, 2x Coarser)\\      ↓\\      Pump V35 (4x Coarser)\end{tabular}} & ARI & {\color[HTML]{FF0000} \textbf{68.04}} & 21.14 & {\color[HTML]{0000FF} {\ul 30.37}} & 28.91 & 29.31 & 4.78 & 14.73 \\ \cline{2-10} 
 &  & ACC & {\color[HTML]{FF0000} \textbf{76.37}} & 5.94 & {\color[HTML]{0000FF} {\ul 9.42}} & 4.43 & 7.59 & 6.05 & 4.56 \\
 &  & MF1 & {\color[HTML]{FF0000} \textbf{33.90}} & 1.98 & {\color[HTML]{0000FF} {\ul 4.70}} & 1.58 & 2.34 & 2.18 & 2.38 \\
 & \multirow{-3}{*}{\begin{tabular}[c]{@{}c@{}}Pump V35 (Original, 4x   Coarser)\\      ↓\\      Pump V35 (2x Coarser)\end{tabular}} & ARI & {\color[HTML]{FF0000} \textbf{64.81}} & 15.02 & {\color[HTML]{0000FF} {\ul 30.86}} & 26.12 & 22.49 & 10.71 & 25.39 \\ \cline{2-10} 
 &  & ACC & {\color[HTML]{FF0000} \textbf{40.27}} & 4.69 & 5.56 & 3.89 & {\color[HTML]{0000FF} {\ul 7.33}} & 4.35 & 3.03 \\
 &  & MF1 & {\color[HTML]{FF0000} \textbf{23.32}} & 1.35 & 1.17 & 1.64 & {\color[HTML]{0000FF} {\ul 2.71}} & 1.32 & 1.79 \\
\multirow{-9}{*}{\begin{tabular}[c]{@{}c@{}}Unseen\\Granularity\\Levels\end{tabular}} & \multirow{-3}{*}{\begin{tabular}[c]{@{}c@{}}Pump V35 (2x, 4x Coarser)\\      ↓\\      Pump V35 (Original)\end{tabular}} & ARI & {\color[HTML]{FF0000} \textbf{31.30}} & 21.37 & 23.40 & 26.91 & 27.45 & -0.41 & {\color[HTML]{0000FF} {\ul 27.64}} \\ \hline
\end{tabular}
}
\end{table*}